\documentclass[11pt]{article}
\usepackage{coling2018}
\usepackage{times}
\usepackage{url}
\usepackage{latexsym}
\usepackage{enumitem}
\usepackage{paralist}

\usepackage{xspace}
\newcommand{\acroft}{FT\xspace}
\newcommand{\acrofsmt}{FSMT\xspace}
\newcommand{\acronmtct}{NMT-constraint\xspace}
\newcommand{\acromttagcat}{MultiTask-tag-style\xspace}
\newcommand{\acromtcat}{MultiTask-style\xspace}
\newcommand{\acromtrandom}{MultiTask-random\xspace}

\newcommand{\acropbmtradom}{PBMT-random\xspace}

\usepackage{graphicx}
\usepackage{subfig}
\usepackage{array}
\usepackage{amssymb}

\title{Multi-Task Neural Models for Translating Between Styles\\ Within and Across Languages}

\author{Xing Niu \\
	University of Maryland \\
	{\tt xingniu@cs.umd.edu} \\\And
	Sudha Rao \\
	University of Maryland \\
	{\tt raosudha@cs.umd.edu} \\\And
	Marine Carpuat \\
	University of Maryland \\
	{\tt marine@cs.umd.edu} \\}

\date{}

\begin{document}
\maketitle
\begin{abstract}
Generating natural language requires conveying content in an appropriate style. We explore two related tasks on generating text of varying formality: monolingual formality transfer and formality-sensitive machine translation. We propose to solve these tasks jointly using multi-task learning, and show that our models achieve state-of-the-art performance for formality transfer and are able to perform formality-sensitive translation without being explicitly trained on style-annotated translation examples. 
\end{abstract}

\section{Introduction}

\blfootnote{
	 \hspace{-0.65cm}  
	 This work is licensed under a Creative Commons 
	 Attribution 4.0 International License.
	 License details:
	 \url{http://creativecommons.org/licenses/by/4.0/}
}

Generating language in the appropriate style is a requirement for applications that generate natural language, as the style of a text conveys important information beyond its literal meaning \cite{hovy1987generating}. \newcite{heylighen1999formality} and \newcite{Biber14} have argued that the formal-informal dimension is a core dimension of stylistic variation. In this work, we focus on the problem of generating text for a desired formality level. 
It has been recently studied in two distinct settings: (1) \newcite{RaoT18} addressed the task of \textit{Formality Transfer} (\acroft) where given an informal sentence in English, systems are asked to output a formal equivalent, or vice-versa; (2) \newcite{NiuMC17} introduced the task of \textit{Formality-Sensitive Machine Translation} (\acrofsmt), where given a sentence in French and a desired formality level (approximating the intended audience of the translation), systems are asked to produce an English translation of the desired formality level. While \acroft and \acrofsmt can both be framed as Machine Translation (MT), appropriate training examples are much harder to obtain than for traditional machine translation tasks. \acroft requires sentence pairs that express the same meaning in two different styles, which rarely occur naturally and are therefore only available in small quantities. \acrofsmt can draw from existing parallel corpora in diverse styles, but would ideally require not only sentence pairs, but e.g., sentence triplets that contain a French input, its formal English translation, and its informal English translation.  

\begin{figure}[h]
	\centering
	\includegraphics[width=1\textwidth]{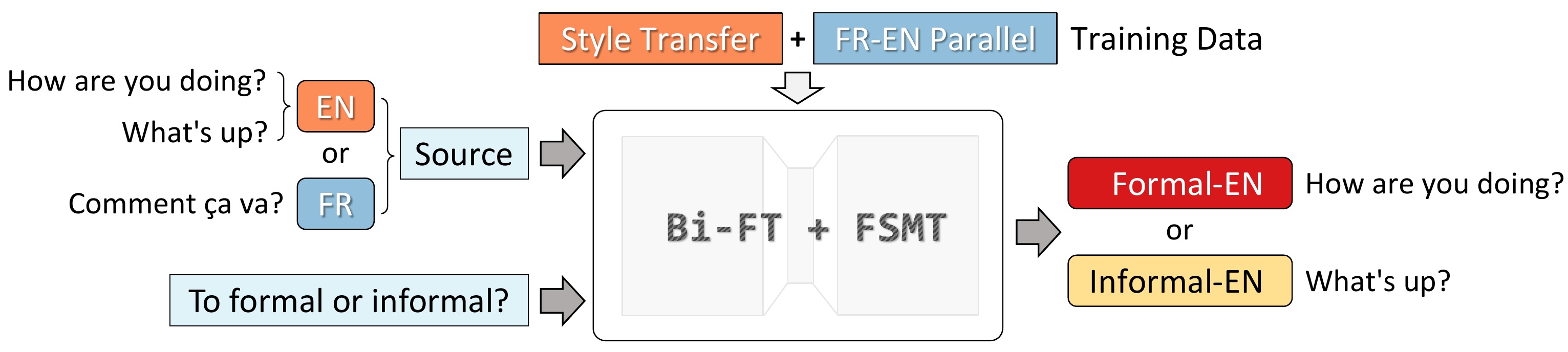}
	\caption{System overview: our multi-task learning model can perform both bi-directional English formality transfer and translate French to English with desired formality. It is trained jointly on monolingual formality transfer data and bilingual translation data.}
	\label{fig:overview}
\end{figure}

We hypothesize that \acroft and \acrofsmt can benefit from being addressed jointly, by sharing information from two distinct types of supervision: sentence pairs in the same language that capture style difference, and translation pairs drawn from corpora of various styles.
Inspired by the benefits of multi-task learning \cite{Caruana97} for natural language processing tasks in general \cite{CollobertW08,LiuGHDDW15,LuongLSVK16}, and for multilingual MT in particular \cite{JohnsonSLKWCTVW17}, we introduce a model based on Neural Machine Translation (NMT) that jointly learns to perform both monolingual \acroft and bilingual \acrofsmt. As can be seen in Figure~\ref{fig:overview}, given an English sentence and a tag (formal or informal), our model paraphrases the input sentence into the desired formality. The same model can also take in a French sentence, and produce a formal or an informal English translation as desired.

Designing this model requires addressing several questions: Can we build a single model that performs formality transfer in both directions?
How to best combine monolingual examples of formality transfer and bilingual examples of translation? What kind of bilingual examples are most useful for the joint task? Can our joint model learn to perform \acrofsmt without being explicitly trained on style-annotated translation examples?
We explore these questions by conducting an empirical study on English \acroft and French-English \acrofsmt, using both automatic and human evaluation. Our results show the benefits of the multi-task learning approach, improving the state-of-the-art on the \acroft task, and yielding competitive performance on \acrofsmt without style-annotated translation examples. Along the way, we also improve over prior results on \acroft using a single NMT model that can transfer between styles in both directions.

\section{Background}

\noindent\textbf{Style Transfer} can naturally be framed as a sequence to sequence translation problem given sentence pairs that are paraphrases in two distinct styles. These parallel style corpora are constructed by creatively collecting existing texts of varying styles, and are therefore rare and much smaller than machine translation parallel corpora. For instance, \newcite{XuRDGC12} scrape modern translations of Shakespeare's plays and use a phrase-based MT (PBMT) system to paraphrase Shakespearean English into/from modern English. \newcite{JhamtaniGHN17} improve performance on this dataset using neural translation model with pointers to enable copy actions. The availability of parallel standard and simple Wikipedia (and sometimes additional human rewrites) makes text simplification a popular style transfer task, typically addressed using machine translation models ranging from syntax-based MT \cite{ZhuBG10,XuNPCC16}, phrase-based MT \cite{CosterK11,WubbenBK12} to neural MT \cite{WangCRQ16} trained via reinforcement learning \cite{ZhangL17}.

Naturally occurring examples of parallel formal-informal sentences are harder to find. Prior work relied on synthetic examples generated based on lists of words of known formality \cite{SheikhaI11}. This state of affairs recently changed, with the introduction of the first large scale parallel corpus for formality transfer, GYAFC (Grammarly's Yahoo Answers Formality Corpus). 110K informal sentences were collected from Yahoo Answers and they were rewritten in a formal style via crowd-sourcing, which made it possible to benchmark style transfer systems based on both PBMT and NMT models \cite{RaoT18}. In this work, we leverage this corpus to enable multi-task \acroft and \acrofsmt.

Recent work also explores how to perform style transfer without parallel data. However, this line of work considers transformations that alter the original meaning (e.g., changes in sentiment or topic), while we view style transfer as meaning-preserving. An auto-encoder is used to encode a sequence to a latent representation which is then decoded to get the style transferred output sequence \cite{MuellerGJ17,HuYLSX17,ShenLBJ17,FuTPZY18,PrabhumoyeTSB18}.

\noindent\textbf{Style in Machine Translation} has received little attention in recent MT architectures. \newcite{MimaFI97} improve rule-based MT by using extra-linguistic information such as speaker's role and gender. \newcite{LewisFX15} and \newcite{NiuC16} equate style with domain, and train conversational MT systems by selecting in-domain (i.e. conversation-like) training data. Similarly, \newcite{WintnerMSRP17} and \newcite{MichelN18} take an adaptation approach to personalize MT with gender-specific or speaker-specific data. Other work has focused on specific realizations of stylistic variations, such as T-V pronoun selection for translation into German \cite{SennrichHB16b} or controlling voice \cite{YamagishiKSK16}. In contrast, we adopt the broader range of style variations considered in our prior work, which introduced the \acrofsmt task \cite{NiuMC17}: in \acrofsmt, the MT system takes a desired formality level as an additional input, to represent the target audience of a translation, which human translators implicitly take into account. This task was addressed via $n$-best re-ranking in phrase-based MT --- translation hypotheses whose formality are closer to desired formality are promoted.

\noindent By contrast, in this work we use neural MT which is based on the \textbf{Attentional Recurrent Encoder-Decoder} model \cite{BahdanauCB15,LuongLSVK16}. The input is encoded into a sequence of vector representations while the decoder adaptively computes a weighted sum of these vectors as the context vector for each decoding step.

\noindent In the joint model, we employ \textbf{Side Constraints} as the formality input to restrict the generation of the output sentence  (Figure~\ref{fig:overview}). Prior work has successfully implemented side constraints as a special token added to each source sentence. These tokens are embedded into the source sentence representation and control target sequence generation via the attention mechanism. \newcite{SennrichHB16b} append $<$T$>$ or $<$V$>$ (i.e. T-V pronoun distinction) to the source text to indicate which pronoun is preferred in the German output. \newcite{JohnsonSLKWCTVW17} and \newcite{NiuDC18} concatenate parallel data of various language directions and mark the source with the desired output language to perform multilingual or bi-directional NMT. \newcite{KobusCS17} and \newcite{ChuDK17} add domain tags for domain adaptation in NMT.

\section{Approach}

We describe our unified model for performing \acroft in both directions (Section~\ref{sec:approach-bift}), our \acrofsmt model with side constraints (Section~\ref{sec:approach-sc}) and finally our multi-task learning model that jointly learns to perform \acroft and \acrofsmt (Section~\ref{sec:approach-mtl}). All models rely on the same NMT architecture: attentional recurrent sequence-to-sequence models.

\subsection{Bi-Directional Formality Transfer}\label{sec:approach-bift}

\newcite{RaoT18} used independent neural machine translation models for each formality transfer direction (\texttt{informal$\rightarrow$formal} and \texttt{formal$\rightarrow$informal}).
Inspired by the bi-directional NMT for low-resource languages \cite{NiuDC18}, we propose a unified model that can handle either direction 
--- we concatenate the parallel data from the two directions of formality transfer and attach a tag to the beginning of each source sentence denoting the desired target formality level i.e. $<$\texttt{F}$>$ for transferring to formal and $<$\texttt{I}$>$ for transferring to informal. This enables our \acroft model to learn to transfer to the correct style via attending to the tag in the source embedding.
We train an NMT model on this combined dataset. Since both the source and target sentences come from the same language, we encourage their representations to lie in the same distributional vector space by (1) building a shared Byte-Pair Encoding (BPE) model on source and target data \cite{SennrichHB16a} and (2) tying source and target word embeddings \cite{PressW17}.

\subsection{Formality-Sensitive Machine Translation with Side Constraints}\label{sec:approach-sc}

Inspired by \newcite{SennrichHB16b}, we use side constraints on parallel translation examples to control output formality. At training time, this requires a tag that captures the formality of the target sentence for every sentence pair. Given the vast range of text variations that influence style,  we cannot obtain tags using rules as for T-V pronoun distinctions \cite{SennrichHB16b}. 
Instead, we categorize French-English parallel data into formal vs. informal categories by comparing them to the informal and formal English from the GYAFC corpus.

We adopt a data selection technique, Cross-Entropy Difference (CED) \cite{MooreL10}, to rank English sentences in the bilingual corpus by their relative distance to each style. First, we consider formal English as the target style and define $CED(s)=H_{formal}(s)-H_{informal}(s)$, where $H_{formal}(s)$ is the cross-entropy between a sentence $s$ and the formal language model.
Smaller CED indicates an English sentence that is more similar to the formal English corpus and less similar to the informal English corpus. We rank English sentences by their CED scores and select the top $N$ sentences (choice of N discussed in Section \ref{sec:eval-st}). Pairing these $N$ English sentences with their parallel French source, we get the formal sample of our bilingual data. Similarly, we construct the informal sample using informal English as the target style. Finally, we combine the formal and the informal samples, attach the $<$\texttt{F}$>$ and $<$\texttt{I}$>$ tags to corresponding source French sentences (i.e. the bottom two rows of data in Figure~\ref{fig:td1}) and train an NMT model for our \acrofsmt task.

\subsection{Multi-Task Learning}\label{sec:approach-mtl}

\begin{figure}[t]
	\centering
	\subfloat[Formality tags on bilingual data\newline + 2-Style selection]{
		\label{fig:td1}
		\includegraphics[width=0.32\textwidth]{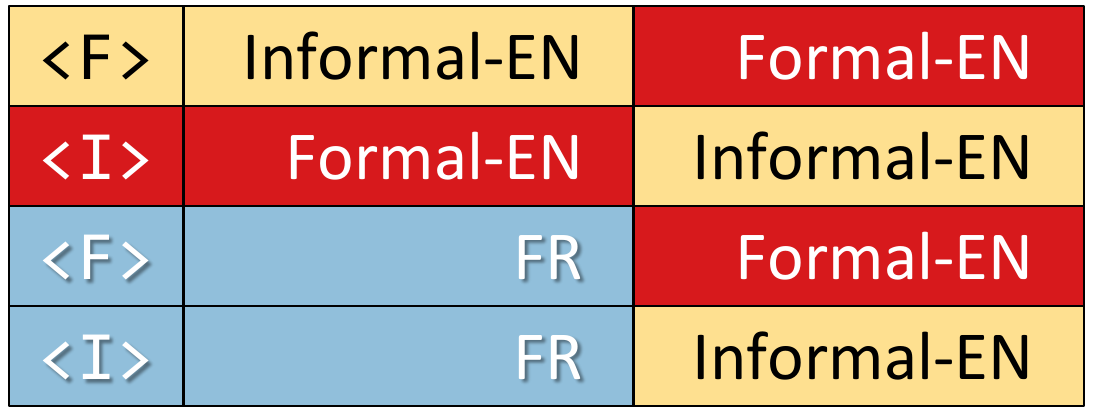}
	}
	\subfloat[No formality tags on bilingual data\newline + 2-Style selection]{
		\label{fig:td2}
		\includegraphics[width=0.32\textwidth]{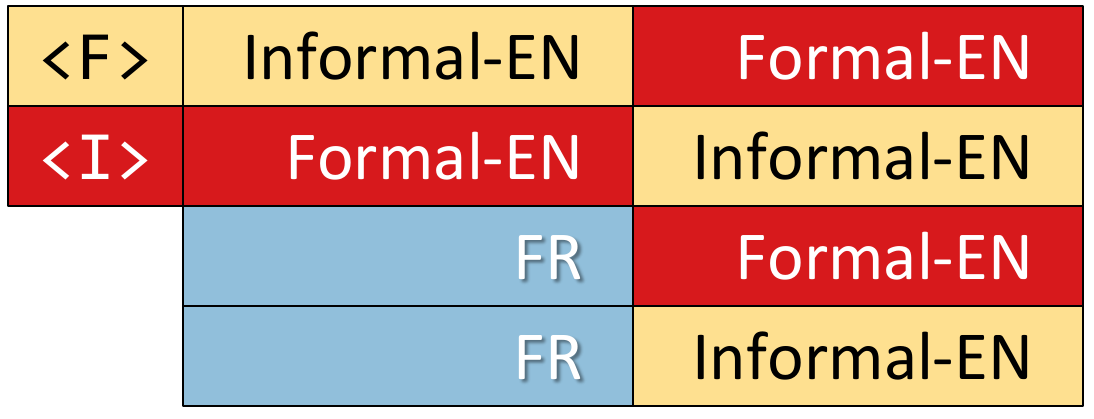}
	}
	\subfloat[No formality tags on bilingual data\newline + Random selection]{
		\label{fig:td3}
		\includegraphics[width=0.32\textwidth]{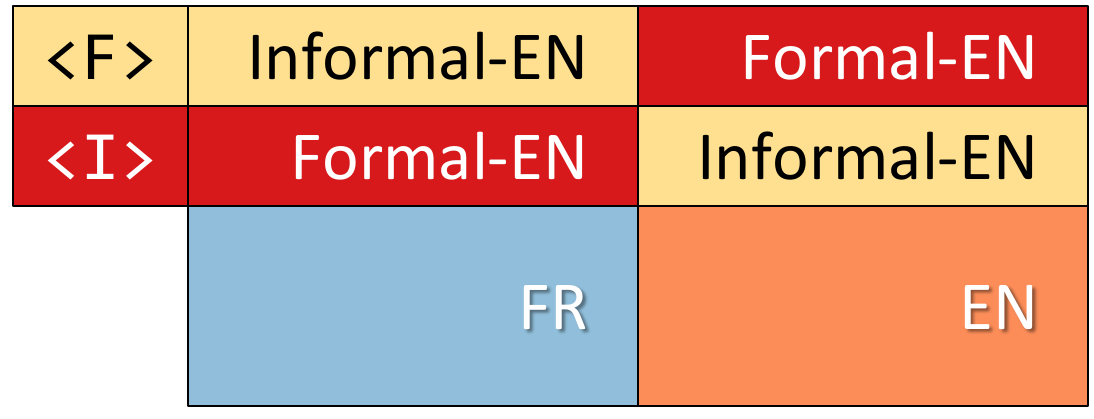}
	}
	\caption{The training data used for multi-task learning models. The bi-directional formality transfer data and the bilingual data (e.g. FR-EN) of equivalent size are always concatenated.}
	\label{fig:training-data}
\end{figure}

We propose a multi-task learning model to jointly perform \acroft and \acrofsmt using a many-to-one (i.e. multi-language to English) sequence to sequence model \cite{LuongLSVK16}. Following \newcite{JohnsonSLKWCTVW17}, we implement this approach using shared encoders and decoders. This approach can use existing NMT architectures without modifications. We design three models to investigate how to best incorporate side constraints at training time, and the benefits of sharing representations for style and language. 

\noindent\textbf{\acromttagcat} is a straightforward combination of the transfer and translation models above. We hypothesize that using the bilingual parallel data where English is the target could enhance English \acroft in terms of target language modeling, especially when the bilingual data has similar topics and styles. We therefore combine equal sizes of formality tagged training data (selected as described in Section~\ref{sec:approach-sc}) from our \acroft and \acrofsmt tasks in this configuration (Figure~\ref{fig:td1}).

\noindent\textbf{\acromtcat} is designed to test whether formality tags for bilingual examples are necessary. We hypothesize that the knowledge of controlling target formality for the \acrofsmt task can be learned from the \acroft data since the source embeddings of formality tags are shared between the \acroft and the \acrofsmt tasks. We therefore combine the formality tagged \acroft data with the MT data without their tags (Figure~\ref{fig:td2}).

\noindent\textbf{\acromtrandom} investigates the impact of the similarity between formality transfer and bilingual examples. Selecting bilingual data which is similar to the GYAFC corpus is not necessarily beneficial for the \acrofsmt task especially when French-English bilingual examples are drawn from a domain distant from the GYAFC corpus. In this configuration, we test how well our model performs \acrofsmt if bilingual examples are randomly selected instead (Figure~\ref{fig:td3}).

\section{Experimental Setup}

\noindent\textbf{\acroft data:} We use the GYAFC corpus introduced by \newcite{RaoT18} as our \acroft data. This corpus consists of 110K informal sentences from two domains of Yahoo Answers (\textit{Entertainment and Music (E\&M)} and \textit{Family and Relationships (F\&R)}) paired with their formal rewrites by humans. The train split consists of 100K informal-formal sentence pairs whereas the dev/test sets consist of roughly 5K source-style sentences paired with four reference target-style human rewrites for both transfer directions.

\noindent\textbf{\acrofsmt data:}  We evaluate the \acrofsmt models on a large-scale French to English (FR-EN) translation task. Examples are drawn from OpenSubtitles2016 \cite{LisonT16} which consists of movie and television subtitles and is thus more similar to the GYAFC corpus compared to news or parliament proceedings. This is a noisy dataset where aligned French and English sentences often do not have the same meaning, so we use a bilingual semantic similarity detector to select 20,005,000 least divergent examples from $\sim$27.5M deduplicated sentence pairs in the original set \cite{VyasNC18}. Selected examples are then randomly split into a 20M training pool, a 2.5K dev set and a 2.5K test set.

\noindent\textbf{Preprocessing:} We apply four pre-processing steps to both \acroft and MT data: normalization, tokenization, true-casing, and joint source-target BPE with 32,000 operations for NMT \cite{SennrichHB16a}.

\noindent\textbf{NMT Configuration:} We use the standard attentional encoder-decoder architecture implemented in the Sockeye toolkit \cite{CoRR:Sockeye}. Our translation model uses a bi-directional encoder with a single LSTM layer \cite{BahdanauCB15} of size 512, multilayer perceptron attention with a layer size of 512, and word representations of size 512.
We apply layer normalization and tie the source and target embeddings as well as the output layer's weight matrix.
We add dropout to embeddings and RNNs of the encoder and decoder with probability 0.2.
We train using the Adam optimizer with a batch size of 64 sentences and checkpoint the model every 1000 updates \cite{KingmaB15}. Training stops after 8 checkpoints without improvement of validation perplexity. We decode with a beam size of 5.
We train four randomly seeded models for each experiment and combine them in a linear ensemble for decoding.\footnote{Data and scripts available at \url{https://github.com/xingniu/multitask-ft-fsmt}.}

\section{Evaluation Protocol}

\subsection{Automatic Evaluation}

We evaluate both \acroft and \acrofsmt tasks using BLEU \cite{PapineniRWZ02}, which compares the model output with four reference target-style rewrites for \acroft and a single reference translation for \acrofsmt. We report case-sensitive BLEU with standard WMT tokenization.\footnote{\url{https://github.com/EdinburghNLP/nematus/blob/master/data/multi-bleu-detok.perl}} For \acroft, \newcite{RaoT18} show that BLEU correlates well with the overall system ranking assigned by humans. For \acrofsmt, BLEU is an imperfect metric as it conflates mismatches due to translation errors and due to correct style variations. We therefore turn to human evaluation to isolate formality differences from translation quality.

\subsection{Human Evaluation}

Following \newcite{RaoT18}, we assess model outputs on three criteria: \textit{formality}, \textit{fluency} and \textit{meaning preservation}. Since the goal of our evaluation is to compare models, our evaluation scheme asks workers to compare sentence pairs on these three criteria instead of rating each sentence in isolation. We collect human judgments using CrowdFlower 
on 300 samples of each model outputs. For \acroft, we compare the top performing NMT benchmark model in \newcite{RaoT18} with our best \acroft model. For \acrofsmt, we compare outputs from three representative models: \acronmtct, \acromtrandom and \acropbmtradom.\footnote{Note that we do not compare with the English reference translation. A more detailed description of the human annotation protocol can be found in the appendix.}

\noindent\textbf{Formality.} For \acroft, we want to measure the amount of style variation introduced by a model. Hence, we ask workers to compare the source-style sentence with its target-style model output. For \acrofsmt, we want to measure the amount of style variation between two different translations by the same model. Hence, we ask workers to compare the ``informal'' English translation and the ``formal'' English translation of the same source sentence in French.\footnote{Evaluating which systems produces the most (in)formal output is an orthogonal question that we leave to future work.} We design a five point scale for comparing the formality of two sentences ranging from one being much more formal than the other to the other being much more formal than the first, giving us a value between 0 and 2 for each sentence pair.\footnote{Details on the conversion from a five point scale to a value between 0 and 2 is in the appendix.}

\noindent\textbf{Fluency.} For both \acroft and \acrofsmt tasks, we want to understand how fluent are the different model outputs. Hence, we ask workers to compare the fluency of two model outputs of the same target style. Similar to formality evaluation, we design a five point scale for comparing the fluency of two sentences, giving us a value between 0 and 2 for each sentence pair.

\noindent\textbf{Meaning Preservation.} For \acroft, we want to measure the amount of meaning preserved during formality transfer. Hence, we ask workers to compare the source-style sentence and the target-style model output. For \acrofsmt, we want to measure the amount of meaning preserved between two different translations by the same model. Hence, we ask workers to compare the ``informal'' English translation and the ``formal'' English translation of the same source sentence in French. We design a four point scale to compare the meaning of two sentences ranging from the two being completely equivalent to the two being not equivalent, giving us a value between 0 and 3 for each sentence pair.

\section{Formality Transfer Experiments}
\label{sec:eval-st}

\subsection{Baseline Models from \newcite{RaoT18}}
\noindent\textbf{PBMT} is a phrase-based machine translation model trained on the GYAFC corpus using a training regime consisting of self-training, data sub-selection and a large language model.

\noindent\textbf{NMT Baseline} uses OpenNMT-py \cite{KleinKDSR17}. \newcite{RaoT18} use a pre-processing step to make source informal sentences more formal and source formal sentences more informal by rules such as re-casing. Word embeddings pre-trained on Yahoo Answers are also used.

\noindent\textbf{NMT Combined} is Rao and Tetreault's best performing NMT model trained on the rule-processed GYAFC corpus, with additional forward and backward translations produced by the PBMT model.

\subsection{Our Models}

\noindent\textbf{NMT Baseline}: Our NMT baseline uses Sockeye instead of OpenNMT-py and is trained on raw datasets of two domains and two transfer directions.

\noindent\textbf{Bi-directional \acroft}: Our initial bi-directional model is trained on bi-directional data from both domains with formality tags. It is incrementally augmented with three modifications to get the final multi-task model (i.e. \acromttagcat as described in Section \ref{sec:approach-mtl}): (1) We combine training sets of two domains (E\&M+F\&R) together and train a single model on it. (2) We use ensemble decoding by training four randomly seeded models on the combined data. (3) We add formality-tagged bilingual data and train the model using multi-task learning to jointly learn \acroft and \acrofsmt. Suppose the amount of original bi-directional \acroft data is $n$, we always select $kn$ bilingual data where $k$ is an integer. We also duplicate \acroft data to make it match the size of selected bilingual data.

\subsection{Results}
\label{sec:ft-results}

\begin{table*}[t]
	\begin{center}
		\scalebox{1}{
			\begin{tabular}{l|>{\centering}p{48pt}c|>{\centering}p{48pt}c}
				& \multicolumn{2}{c|}{\texttt{Informal$\rightarrow$Formal}} & \multicolumn{2}{c}{\texttt{Formal$\rightarrow$Informal}} \\
				Model & E\&M & F\&R & E\&M & F\&R \\
				\hline
				PBMT \cite{RaoT18} & 68.22 & 72.94 & 33.54 & 32.64 \\
				NMT Baseline \cite{RaoT18} & 58.80 & 68.28 & 30.57 & 36.71 \\
				NMT Combined \cite{RaoT18} & 68.41 & 74.22 & 33.56 & 35.03 \\
				\hline
				NMT Baseline & 65.34 & 71.28 & 32.36 & 36.23 \\
				Bi-directional \acroft & 66.30 & 71.97 & 34.00 & 36.33 \\
				\hspace{6pt} + training on E\&M + F\&R & 69.20 & 73.52 & 35.44 & 37.72 \\
				\hspace{6pt} + ensemble decoding ($\times$4) & 71.36 & 74.49 & 36.18 & 38.34 \\
				\hspace{6pt} + multi-task learning (\acromttagcat) & \bf 72.13 & \bf 75.37 & \bf 38.04 & \bf 39.09 \\
		\end{tabular}}
	\end{center}
	\caption{Automatic evaluation of Formality Transfer with BLEU scores. The bi-directional model with three stacked improvements achieves the best overall performance. The improvement over the second best system is statistically significant at $p<0.05$ using bootstrap resampling \cite{Koehn04}.}
	\label{tab:st}
\end{table*}

\begin{table}[t]
	\begin{center}
    \scalebox{0.91}{
    	\begin{tabular}{c|l|l|cc|c}
			& & & \multicolumn{2}{c|}{Formality Diff} & Meaning Preservation \\
            & Model A & Model B & \multicolumn{2}{c|}{Range = [0,2]} & Range = [0,3] \\
            \hline
            & & & \texttt{I$\rightarrow$F} & \texttt{F$\rightarrow$I} & \\
            \cline{4-5}
            \acroft & Source & NMT Combined & 0.54 & 0.45 & 2.94\\
            & Source & \acromttagcat & 0.59 & \bf 0.64 & 2.92 \\
            \hline
            & \acronmtct \texttt{I} & \acronmtct \texttt{F} & \multicolumn{2}{c|}{\bf 0.35} & 2.95\\
            \acrofsmt & NMT \acromtrandom \texttt{I} & NMT \acromtrandom \texttt{F} & \multicolumn{2}{c|}{\bf 0.32} & 2.90\\
            & \acropbmtradom \texttt{I} & \acropbmtradom \texttt{F} & \multicolumn{2}{c|}{0.05} & 2.97\\
		\end{tabular}}
	\end{center}
	\caption{Human evaluation of formality difference and meaning preservation. \acromttagcat generates significantly more informal (F$\rightarrow$I) English than NMT Combined ($p$$<$0.05 using the t-test, see Section~\ref{sec:ft-results}). \acropbmtradom does not control formality effectively when comparing its informal (I) and formal (F) output (Section~\ref{sec:fsmt-results}). Formality scores are relatively low because workers rarely choose ``much more (in)formal''. All models preserve meaning equally well.}
	\label{tab:human-eval}
\end{table}

\begin{figure}[t]
	\centering
	\subfloat[BLEU improvements for formality transfer.]{
		\label{fig:name1}
		\includegraphics[width=0.49\textwidth]{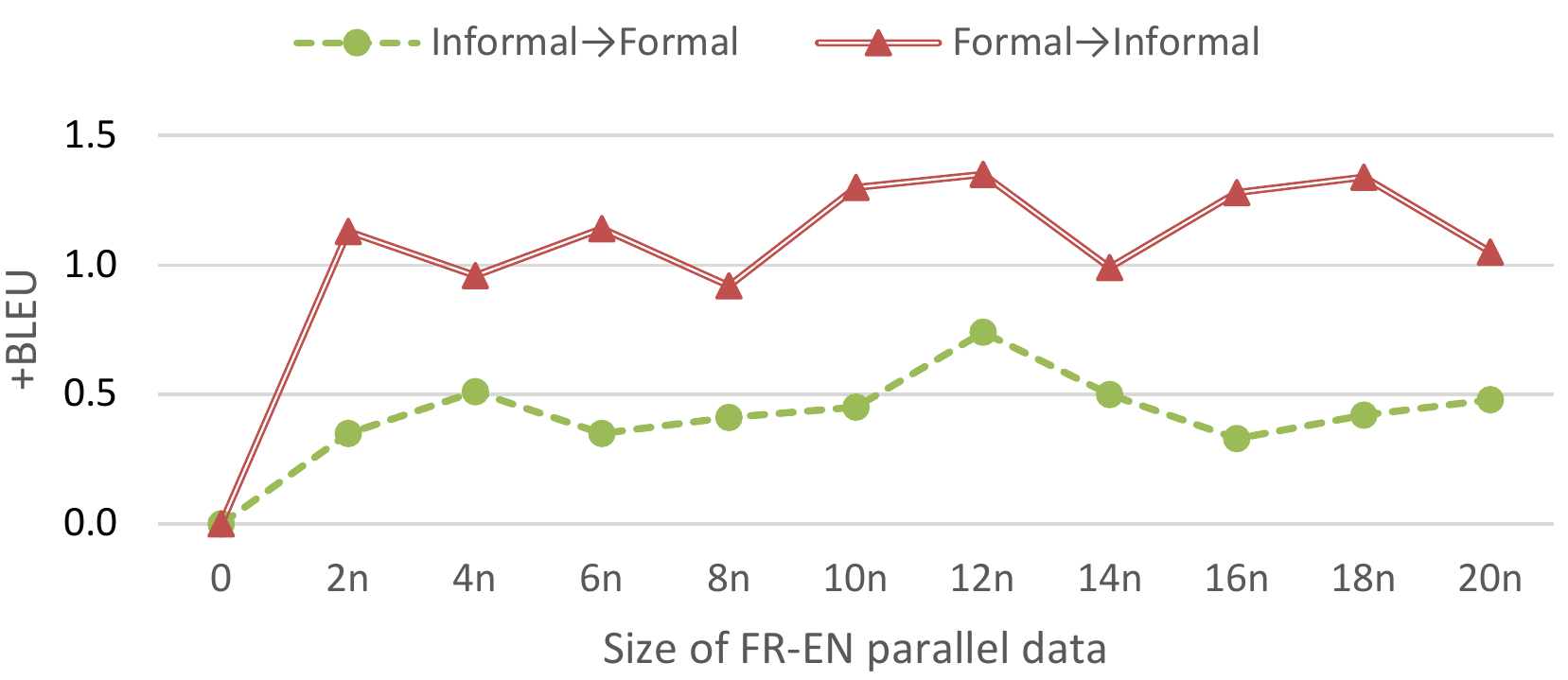}
	}
	\subfloat[BLEU scores for machine translation.]{
		\label{fig:name2}
		\includegraphics[width=0.49\textwidth]{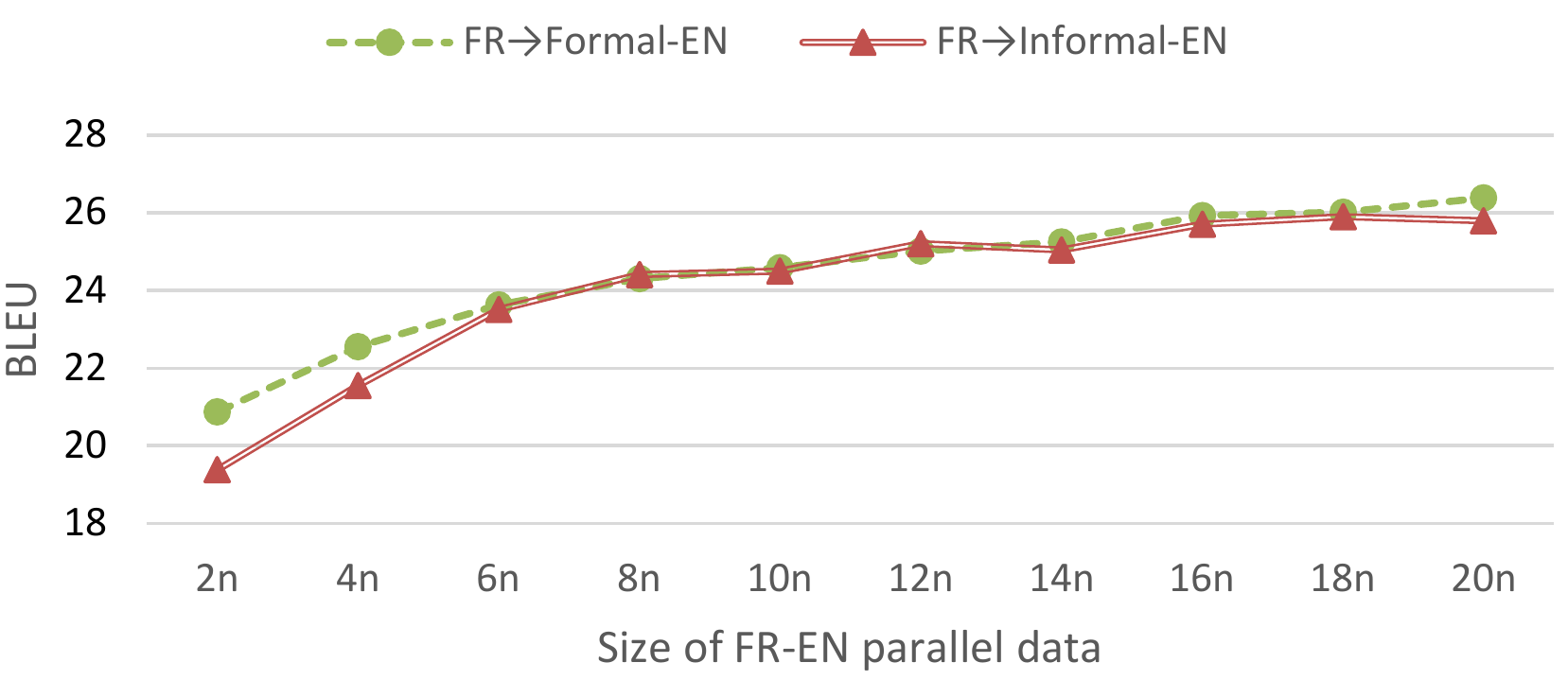}
	}
	\caption{BLEU improvements or scores for four transfer/translation directions vs. the size of FR-EN parallel data. $n$ in x-axis equals to the original size of bi-directional style transfer training data. Formality transfer improves with bilingual data and the performance converges quickly. The translation quality increases monotonously with the size of training data.}
	\label{fig:st-mt}
\end{figure}

\noindent\textbf{Automatic Evaluation.} As shown in Table~\ref{tab:st}, our NMT baselines yield surprisingly better BLEU scores than those of \newcite{RaoT18}, even without using rule-processed source training data and pre-trained word embeddings. We attribute the difference to the more optimized NMT toolkit we use.

Initial bi-directional models outperforms uni-directional models. This matches the behavior of bi-directional NMT in low-resource settings studied by \newcite{NiuDC18} --- we work with a relatively small amount of training data ($\sim$50K), and \acroft models benefit from doubling the size of training data without being confused by mixing two transfer directions. For the same reason, increasing the training data by combining two domains together improves performance further. Ensemble decoding is a consistently effective technique used by NMT and it enhances our NMT-based \acroft models as expected.

Incorporating the bilingual parallel data by multi-task learning yields further improvement. The target side of bilingual data is selected based on the closeness to the GYAFC corpus, so we hypothesize that the higher quality comes from better target language modeling by training on more English text.

\noindent\textbf{Human Evaluation.} The superior performance of the best \acroft model (i.e. \acromttagcat) is also reflected in our human evaluation (see Table~\ref{tab:human-eval}). It generates slightly more formal English (0.59 vs 0.54) and significantly more informal English (0.64 vs 0.45) than NMT Combined. This is consistent with BLEU differences in Table~\ref{tab:st} which show that \acromttagcat yields bigger improvements when transferring formal language to informal. Both models have good quality with respect to meaning preservation (2.94 vs 2.92) and workers can hardly find any fluency difference between outputs of these two models by assigning 0.03 in average in the fluency test (0 means no difference).

\noindent\textbf{Impact of Bilingual Data Size.} We evaluate the impact of selected bilingual data size on the combination of development sets from two domains in GYAFC and show the results in Figure~\ref{fig:st-mt}. The quality of formality transfer improves instantly when using bilingual data and it soon converges when more data is used. Meanwhile, the translation quality increases monotonously with the size of training data. The optimal point is a hyper-parameter that can be determined on the development set. We empirically choose $n=12$ since it works best for formality transfer and yields reasonable translation quality.

\begin{table}[t]
\scalebox{0.83}{
\begin{tabular}{l l l}
\hline
\multicolumn{3}{l}{\textbf{~\ref{tab:model_outputs_st}a: \texttt{informal$\rightarrow$formal}}}\\
\hline
 & Original \texttt{I} & chill out sweetie everything will be just fine eventually\\
1 & NMT-Combined \texttt{F}  & Can you chill out sweetie everything will be just fine eventually.\\
& \acromttagcat \texttt{F}  & Calm down, sweetie, everything will be fine eventually.\\
\hline
 & Original \texttt{I}  & Dakota Fanning.....I know that she is only 12 but she is really famous.\\
2 & NMT-Combined \texttt{F} & Dakota Fanning.i know that she is only twelve, but she is famous.\\
& \acromttagcat \texttt{F} & I know that Dakota Fanning is only twelve, but she is really famous.\\
\hline
 & Original \texttt{I} & depends....usully they are about ur personailty but not wat ur gonna do iwith ur life.\\
3 & NMT-Combined \texttt{F} & Depends.usully they are about your personailty, but not what your going to do iwith your life.\\
& \acromttagcat \texttt{F} & It depends. They are about your personality, but not what you are going to do with your life.\\
\hline
 & Original \texttt{I}  & THAT DEPENDS...ARE YOU A HOTTIE W/A BODY?\\
4 & NMT-Combined \texttt{F} & That depends, are you a hottie with a body?\\
& \acromttagcat \texttt{F} & That depends. Are you a HOTTIE W / A BODY?\\
\hline
\multicolumn{3}{l}{\textbf{~\ref{tab:model_outputs_st}b: \texttt{formal$\rightarrow$informal}}}\\
\hline
 & Original \texttt{F} & Therefore I would say that they do succeed but not frequently. I hope this is helpful.\\
1 & NMT-Combined \texttt{I}  & So I would say that they do failing but not frequently, I hope this is helps.\\
& \acromttagcat \texttt{I}  & so i would say they do it but not all the time, hope this helps.\\
\hline
& Original \texttt{F} & I am simply inquiring because people behave as though they are no longer interested in them.\\
2 & NMT-Combined \texttt{I}  & I am just asking because people act as though they are no longer interested in them.\\
& \acromttagcat \texttt{I}  & I'm just asking because people act like they don't like them anymore.\\
\hline
 & Original \texttt{F} & Hello, I am interested in visiting your country.\\
3 & NMT-Combined \texttt{I} & Hi, I'm interested in visiting your country.\\
& \acromttagcat \texttt{I} & hi, I'm going to go to your country.\\
\hline
\end{tabular}}
\caption{Sample model outputs for the Formality Transfer (\acroft) task.}
\label{tab:model_outputs_st}
\end{table}

\subsection{Qualitative Analysis}
We manually inspect 100 randomly selected samples from our evaluation set and compare the target-style output of our best model (\acromttagcat) with that of the best baseline model (NMT-Combined) from \newcite{RaoT18}. Table~\ref{tab:model_outputs_st} shows some samples representative of the trends we find for \texttt{informal$\rightarrow$formal} (\ref{tab:model_outputs_st}a) and \texttt{formal$\rightarrow$informal} (\ref{tab:model_outputs_st}b) tasks.

In majority of the cases, the two models produce similar outputs as can be expected since they use similar NMT architectures. In cases where the two outputs differ, in the \texttt{I$\rightarrow$F} task, we find that our model produces a more formal output by introducing phrasal level changes (first sample in~\ref{tab:model_outputs_st}a) or by moving phrases around (second sample in~\ref{tab:model_outputs_st}a), both of which happens frequently during machine translation thus showcasing the benefit of our multi-task approach. Our model very often makes the output sentence more complete (and thereby more formal) by inserting pronouns like `it', `they' at the start of the sentence or by removing conjunctions like `usually', `and', `but', `however' from the beginning of a sentence (sample three in~\ref{tab:model_outputs_st}a). Likewise, in the \texttt{F$\rightarrow$I} task, our model produces more informal sentences compared to the baseline by introducing more phrasal level changes (first and second sample in~\ref{tab:model_outputs_st}b).

\paragraph{Error analysis:} In the \texttt{I$\rightarrow$F} task, our model performs worse than the baseline when the original informal sentence consists of all uppercased words (fourth sample in~\ref{tab:model_outputs_st}a). This is primarily because the baseline model pre-lowercases them using rules. Whereas, we rely on the model to learn this transformation and so it fails to do so for less frequent words. In the \texttt{F$\rightarrow$I} task, in trying to produce more informal outputs, our model sometimes fails to preserve the original meaning of the sentence (third sample in~\ref{tab:model_outputs_st}b). In both tasks, very often our model fails to make transformations for some pairs like (`girls',`women'), 
which the baseline model is very good at. We hypothesize that this could be because for these pairs, human rewriters do not always agree on one of the words in the pair being more informal/formal. This makes our model more conservative in making changes because our bi-directional model combines \acroft data from both directions and when the original data contains instances where these words are not changed, we double that and learn to copy the word more often than change it.

\section{Formality-Sensitive Machine Translation Experiments}

\begin{table*}[t]
	\begin{center}
		\scalebox{0.9}{
			\begin{tabular}{l|c|c|c|c}
				Model & +Tag? & Random? & \texttt{FR$\rightarrow$Formal-EN} & \texttt{FR$\rightarrow$Informal-EN} \\
				\hline
                \acronmtct & $\checkmark$ & & 27.15 & 26.70 \\
				NMT \acromttagcat & $\checkmark$ &  & 25.02 & 25.20 \\
				NMT \acromtcat & & & 23.25 & 23.41 \\
				NMT \acromtrandom & & $\checkmark$ & 25.24 & 25.14 \\
				\acropbmtradom (Niu et al.) & & $\checkmark$ & 29.12 & 29.02 \\
		\end{tabular}}
	\end{center}
	\caption{BLEU scores of various \acrofsmt models. ``+Tag" indicates using formality tags for bilingual data while ``Random" indicates using randomly selected bilingual data.}
	\label{tab:mt}
\end{table*}

\subsection{Models}

\noindent\textbf{\acronmtct}: We first evaluate the standard NMT model with side constraints introduced in Section \ref{sec:approach-sc} and then compare it with three variants of \acrofsmt models using multi-task learning as described in Section \ref{sec:approach-mtl} (i.e. \textbf{\acromttagcat}, \textbf{\acromtcat} and \textbf{\acromtrandom}). The best performing system for \acroft is \acromttagcat with 12$n$ ($\sim$2.5M) bilingual data. For fair comparison, we select this size of bilingual data for all \acrofsmt models either by data selection or randomly.

\noindent\textbf{\acropbmtradom}: We also compare our models with the PBMT-based \acrofsmt system proposed by \newcite{NiuMC17}. Instead of tagging sentences in a binary fashion, this system scores each sentence using a lexical formality model. It requests a desired formality score for translation output and re-ranks $n$-best translation hypotheses by their closeness to the desired formality level. We adapt this system to our evaluation scenario --- we calculate median scores for informal and formal data (i.e. $-0.41$ and $-0.27$ respectively) in GYAFC respectively by a PCA-LSA-based formality model \cite{NiuC17,NiuMC17}
and use them as desired formality levels.\footnote{The PCA-LSA-based formality model achieves lowest root-mean-square error on a scoring task of sentential formality as listed on \url{https://github.com/xingniu/computational-stylistic-variations}.}
The bilingual training data is randomly selected.

\subsection{Results}
\label{sec:fsmt-results}

\noindent\textbf{Automatic Evaluation.} We compute BLEU scores on the held out test set for all models as a sanity check on translation quality. Because there is only one reference translation of unknown style for each input sentence, these BLEU scores conflate translation errors and stylistic mismatch, and are therefore not sufficient to evaluate \acrofsmt performance. We include them for completeness here, as indicators of general translation quality, and will rely on human evaluation as primary evaluation method. As can be seen in Table~\ref{tab:mt}, changing the formality level for a given system yields only small differences in BLEU. Based on BLEU scores, we select \acromtrandom as the representative of multi-task \acrofsmt and compare it with \acronmtct and \acropbmtradom during our human evaluation.

\noindent\textbf{Human Evaluation.} Table~\ref{tab:human-eval} shows that neural models control formality significantly better than \acropbmtradom (0.35/0.32 vs. 0.05). They also introduce more changes in translation: with NMT models, $\sim$80\% of outputs change when only the input formality changes, while that is only the case for $\sim$30\% of outputs with \acropbmtradom. Among neural models, \acromtrandom and \acronmtct have similar quality in controlling output formality (0.32 vs. 0.35) and preserving meaning (2.90 vs. 2.95). They are also equally fluent as judged by humans. Interestingly, multi-task learning helps \acromtrandom perform as well as \acronmtct with simpler examples that do not require the additional step of data selection to generate formality tags.

\subsection{Qualitative Analysis}

We randomly sample 100 examples from our test set and manually compare the formal and the informal translations of the French source by \acromtrandom, \acronmtct and \acropbmtradom. Table~\ref{tab:model_outputs_fsmt} shows representative examples of the observed trends.

We find that in most cases, the difference between the formal and informal style translations is very minor in \acropbmtradom model, better in \acronmtct model and the best in our \acromtrandom model (first sample in the table). In general, our \acromtrandom model does a good job of making very large changes while transferring the style, especially into informal (second sample in the table). We hypothesize that this is because our joint model is trained on the GYAFC corpus which consists of parallel sentences that differ heavily in style.

\paragraph{Error analysis:} All \acrofsmt models perform well in terms of meaning preservation, yet the human scores are not perfect (Table~\ref{tab:human-eval}). They occasionally change not only the style but also the meaning of the input (e.g. the third sample of \acromtrandom in Table~\ref{tab:model_outputs_fsmt}). This motivates future work that penalizes meaning changes more explicitly during training. In general, none of the models do a good job of changing the style when the source sentence is not skewed in one style. For example, consider the French sentence ``Combien de fois vous l'ai-je dit?'' and its English reference translation ``How many times have I told you, right?''. All models produce the same translation ``How many times did I tell you?''. In such cases, changing style requires heavier editing or paraphrasing of the source sentence that our current models are unable to produce.

\begin{table}[t]
	\scalebox{0.87}{
		\begin{tabular}{l l l}
			\hline
			1 & French Source & Impossible d'avoir acc\`es \`a internet ici.\\
			& English Reference & I don't know if you've tried yet, but it's impossible to get on the internet up here.\\
			\hline
			& \acromtrandom Formal & It is impossible to have access to the internet here. \\
			& \acronmtct Formal & It's impossible to have access to the Internet here. \\
			& \acropbmtradom Formal & I can't access to the internet here. \\
			\hline
			& \acromtrandom Informal & Impossible to get to the internet here. \\
			& \acronmtct Informal & Couldn't have accessed the internet here.\\
			& \acropbmtradom Informal & I can't access to the internet here.\\
			\hline
			\hline
			2 & French Source & Abstenez-vous de tout commentaire et r\'epondez \`a la question, chef Toohey. \\
			& English Reference & Refrain from the commentary and respond to the question, Chief Toohey.\\
			\hline
			& \acromtrandom Formal & You need to be quiet and answer the question, Chief Toohey. \\
			& \acronmtct Formal & Please refrain from any comment and answer the question, Chief Toohey.\\
			& \acropbmtradom Formal & Please refrain from comment and just answer the question, the Tooheys's boss.\\
			\hline
			& \acromtrandom Informal & Shut up and answer the question, Chief Toohey.\\
			& \acronmtct Informal & Please refrain from comment and answer the question, chief Toohey.\\
			& \acropbmtradom Informal & Please refrain from comment and answer my question, Tooheys's boss.\\
			\hline
			\hline
			3 & French Source &  Essaie de pr\'esenter des requ\^etes suppl\'ementaires d\`es que tu peux. \\
			& English Reference & Try to file any additional motions as soon as you can. \\
			\hline
			& \acromtrandom Formal & You should try to introduce the sharks as soon as you can.\\
			& \acronmtct Formal & Try to present additional requests as soon as you can. \\
			& \acropbmtradom Formal & Try to introduce any additional requests as soon as you can.\\
			\hline
			& \acromtrandom Informal & Try to introduce sharks as soon as you can.\\
			& \acronmtct Informal & Try to introduce extra requests as soon as you can.\\
			& \acropbmtradom Informal & Try to introduce any additional requests as soon as you can. \\
			\hline
	\end{tabular}}
	\caption{Sample model outputs for the Formality-Sensitive Machine Translation (FSMT) task.}\label{tab:model_outputs_fsmt}
\end{table}

\section{Conclusion}

We explored the use of multi-task learning to jointly perform monolingual \acroft and bilingual \acrofsmt. Using French-English translation and English style transfer data, we showed that the joint model is able to learn from both style transfer parallel examples and translation parallel examples. On the \acroft task, the joint model significantly improves the quality of transfer between formal and informal styles in both directions, compared to prior work \cite{RaoT18}. The joint model interestingly also learns to perform \acrofsmt without being explicitly trained on style-annotated translation examples. On the \acrofsmt task, our model outperforms previously proposed phrase-based MT model \cite{NiuMC17}, and performs on par with a neural model with side-constraints which requires more involved data selection.

These results show the promise of multi-task learning for controlling style in language generation applications. In future work, we plan to investigate other multi-task architectures and objective functions that better capture the desired output properties, in order to help address current weaknesses such as meaning errors revealed by manual analysis.

\section*{Acknowledgments}

We thank the three anonymous reviewers for their helpful comments and suggestions. We thank Joel Tetreault for useful discussions and for making the GYAFC corpus available, as well as members of the Computational Linguistics and Information Processing (CLIP) lab at University of Maryland for helpful discussions. This work is supported by the Clare Boothe Luce Foundation and by the NSF grant IIS-1618193. Any opinions, findings, conclusions, or recommendations expressed here are those of the authors and do not necessarily reflect the view of the sponsors.

\bibliographystyle{acl}
\bibliography{stytrans-nmt}

\newpage
\section*{Appendix A. Details of Human-Based Evaluations}\label{human-eval}
As described in the main paper, we assess the model outputs on three criteria of formality, fluency and meaning preservation. We collect these judgments using CrowdFlower. 

Since we want native English speakers to perform this task, we restrict our set of annotators only to these three native English speaking countries: United States, United Kingdom and Australia. We create a sample of 51 gold questions for each of the three tasks (criteria). Annotators have to continually maintain an accuracy of above 70\% to be able to contribute to the task.

We collect judgments on 300 samples of each model output and we collect three judgments per sample (i.e. sentence pair). Given the three judgments per sample, we calculate the aggregate score using the weighted average:
\[\frac{\sum_{i=1}^{3} score_i \times trust_i}{\sum_{i=1}^{3} trust_i}\]
where $score_i$ is the score given by an annotator and $trust_i$ is our  trust on that annotator. This trust is the accuracy of the annotator on the gold questions.

\paragraph{Formality:} Given two sentences, we ask workers to compare their formality using one of the following categories, regardless of fluency and meaning. We do not enumerate specific rules (e.g. typos or contractions) and encourage workers to use their own judgment.

\begin{table}[h]
\begin{tabular}{c | l}
\textbf{Score} & \textbf{Category} \\
\hline
2 & \textit{Sentence 1 is much more formal than Sentence 2} \\
1 & \textit{Sentence 1 is more formal than Sentence 2} \\
0 & \textit{No difference or hard to say} \\
-1 & \textit{Sentence 2 is more formal than Sentence 1} \\
-2 & \textit{Sentence 2 is much more formal than Sentence 1} \\
\end{tabular}
\end{table}

\paragraph{Fluency:} Given two sentences, we ask workers to compare their fluency using one of the following categories, regardless of style and meaning. We define fluency as follows: \textit{A sentence is fluent if it has a meaning and is coherent and grammatical well-formed.}
\begin{table}[h]
\begin{tabular}{c | l}
\textbf{Score} & \textbf{Category} \\
\hline
2 & \textit{Sentence 1 is much more fluent than Sentence 2} \\
1 & \textit{Sentence 1 is more fluent than Sentence 2} \\
0 & \textit{No difference or hard to say} \\
-1 & \textit{Sentence 2 is more fluent than Sentence 1} \\
-2 & \textit{Sentence 2 is much more fluent than Sentence 1} \\
\end{tabular}
\end{table}

\paragraph{Meaning preservation:} Given two sentences, we ask workers to answer ``how much of the first sentence's meaning is preserved in the second sentence", regardless of style.
\begin{table}[h]
\begin{tabular}{c | l}
\textbf{Score} & \textbf{Category} \\
\hline
3 & \textit{Equivalent since they convey the same key idea}\\
2 & \textit{Mostly equivalent  since they convey the same key idea but differ in some unimportant details}\\
1 & \textit{Roughly equivalent since they share some ideas but differ in important details}\\
0 & \textit{Not equivalent since they convey different ideas}
\end{tabular}
\end{table}

While collecting formality and fluency annotations for sentence pairs, to avoid system-level bias, we randomly swap the two items in the pair and collect annotations on a symmetric range of [-2,2]. But while aggregating these scores, we recover the order and hence the final scores in Table \ref{tab:human-eval} are only in the range of [0,2]. 

\end{document}